\documentclass[sn-mathphys,Numbered]{sn-jnl}

\usepackage{graphicx}%
\usepackage{multirow}%
\usepackage{amsmath,amssymb,amsfonts}%
\usepackage{amsthm}%
\usepackage{mathrsfs}%
\usepackage[title]{appendix}%
\usepackage{xcolor}%
\usepackage{textcomp}%
\usepackage{manyfoot}%
\usepackage{booktabs}%
\usepackage{algorithm}%
\usepackage{algorithmicx}%
\usepackage{algpseudocode}%
\usepackage{listings}%
\usepackage{lscape}
\usepackage{subcaption}
\usepackage{lipsum}
\usepackage{caption}

\usepackage{framed,multirow}
\usepackage{amssymb}
\usepackage{latexsym}
\usepackage[utf8]{inputenc}
\usepackage{amsmath}
\usepackage{url}
\usepackage{xcolor}
\definecolor{newcolor}{rgb}{.8,.349,.1}
\usepackage{hyperref}

\theoremstyle{thmstyleone}%
\theoremstyle{thmstyletwo}%

\theoremstyle{thmstylethree}%

\begin{document}

\title{Saliency-based Video Summarization for Face Anti-spoofing}


\author*[1]{\fnm{Usman} \sur{Muhammad}}
\author*[2]{\fnm{Mourad} \sur{Oussalah}}
\author*[3]{\fnm{Jorma} \sur{Laaksonen}}

\affil*[1,2,3]{\orgdiv{Center for Machine Vision and Signal Analysis}, \orgname{University of Oulu}}

\affil*[1,3]{\orgdiv{Department of Computer Science}, \orgname{Aalto University}, \country{Finland}}


\abstract{
With the growing availability of databases for face presentation attack detection, researchers are increasingly focusing on video-based face anti-spoofing methods that involve hundreds to thousands of images for training the models. However, there is currently no clear consensus on the optimal number of frames in a video to improve face spoofing detection. Inspired by the visual saliency theory, we present a video summarization method for face anti-spoofing detection that aims to enhance the performance and efficiency of deep learning models by leveraging visual saliency. In particular, saliency information is extracted from the differences between the Laplacian and Wiener filter outputs of the source images, enabling identification of the most visually salient regions within each frame. Subsequently, the source images are decomposed into base and detail images, enhancing the representation of the most important information. Weighting maps are then computed based on the saliency information, indicating the importance of each pixel in the image. By linearly combining the base and detail images using the weighting maps, the method fuses the source images to create a single representative image that summarizes the entire video. The key contribution of the proposed method lies in demonstrating how visual saliency can be used as a data-centric approach to improve the performance and efficiency for face presentation attack detection. By focusing on the most salient images or regions within the images, a more representative and diverse training set can be created, potentially leading to more effective models. To validate the method's effectiveness, a simple CNN-RNN deep learning architecture was used, and the experimental results showcased state-of-the-art performance on five challenging face anti-spoofing datasets.}

\keywords{Face Anti-Spoofing, Visual saliency,  Deep Learning, Recurrent Neural Network.}



\maketitle

\section{Introduction}\label{sec1}
Face recognition technology has seen widespread use in various domains, including mobile banking, public security, airports, military and law enforcement \cite{muhammad2022adaptive}. However, a significant concern in face recognition technology is coping with presentation attacks, also known as spoofing attacks or biometric vulnerabilities. Presentation attacks involve attempts to deceive or manipulate the face recognition system by presenting artificial or manipulated facial information, such as printed photos, replay attacks, or 3D mask attacks. The goal of these attacks is to impersonate someone else or gain unauthorized access to sensitive information. As attackers continuously evolve their techniques, it has become crucial for face recognition systems to employ robust face anti-spoofing countermeasures. These countermeasures help ensure the security and reliability of face recognition technology in real-world applications \cite{muhammad2023deep}.\\
\indent Recent studies have indicated that video-based methods \cite{shao2020regularized, wang2020cross, muhammad2023self} may offer superior effectiveness for face presentation attack detection (PAD) compared to image-based methods \cite{boulkenafet2016face, wen2015face, muhammad2019face}. The advantage of video-based approaches lies in the inclusion of temporal information, which provides valuable insights into facial movements, texture changes, and other dynamic characteristics that help differentiate real faces from spoofing attacks. Moreover, spoofing attacks often lack natural facial motion and exhibit inconsistent patterns, such as those seen with textured materials like paper or fabric. These inconsistencies can be more easily detected in a video sequence. By considering a sequence of frames, video-based methods can reduce noise, enhance the signal-to-noise ratio, and improve the overall accuracy of face anti-spoofing systems. However, it is crucial to recognize that the effectiveness of video-based methods can be influenced by various factors, including the quality and resolution of the video, frame rate, camera characteristics, and the specific algorithms and features used for analysis \cite{yu2020fas}.\\
\indent Considering a typical face presentation attack detection (PAD) video, using few frames may not provide sufficient temporal information, limiting the system's ability to effectively detect and differentiate presentation attacks. On the other hand, using an excessive number of frames can lead to increased computational burden, memory requirements, and processing time, without substantial performance gains \cite{muhammad2023deep}. To address this challenge, frame selection approaches have gained increasing attention in the face anti-spoofing literature. These methods aim to identify a subset of frames that effectively represent the video sequences with a single image or a reduced set of frames, while still capturing the temporal changes and motion characteristics that distinguish live faces from static representations. One such approach, presented by Usman {\textit{et al.}} \cite{muhammad2023face}, utilizes a Gaussian weighting function to aggregate frame segments into a single image. This frame aggregation helps the model capture the necessary temporal information for detecting presentation attacks effectively.\\
\indent Another approach to frame selection involves identifying key frames based on specific criteria, such as visual saliency, motion information, or scene changes \cite{li2018can}. For example, some methods use visual saliency information in conjunction with optical flow estimation to extract facial motion features \cite{wang2021silicone}. However, it is worth noting that methods using optical flow can be computationally expensive. Global motion estimation is also critical in compensating for camera motion or detecting and analyzing object motion within a scene. Nonetheless, in some cases, the resulting image may have black borders that provide an artificial cue or artifact, necessitating further preprocessing before training anti-spoofing models \cite{muhammad2022adaptive}. On the other hand, adaptive frame averaging strategies \cite{muhammad2022self, muhammad2022adaptive, muhammad2019face} can offer optimal results in handling temporal variations. However, relying solely on frame averaging can lead to motion blur, which can negatively impact the quality of the spatio-temporal features extracted from the video. The frame averaging result is visualized in source image 2 of Figure 1. As a result, detecting real and attack videos poses a fundamental yet challenging problem in face anti-spoofing. Thus, the frame averaging approach needs to strike a balance between capturing meaningful spatio-temporal information while avoiding motion blur and preserving discriminative features.\\
\begin{figure*}
\centering
\includegraphics[width=0.94\textwidth]{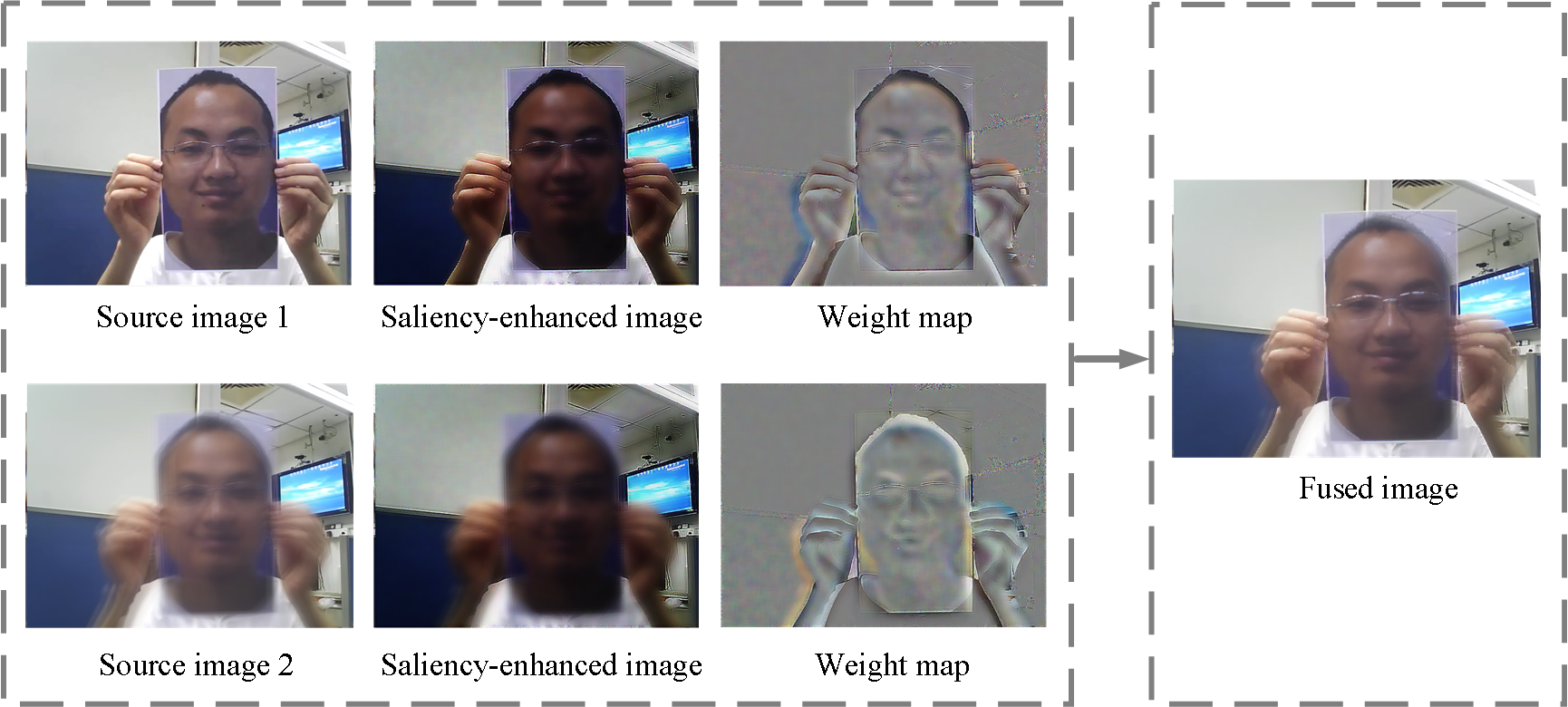}
\vspace{10pt}
\caption{Color-based saliency and weighting maps are used for performing the fusion of two source images.}
\end{figure*}
\indent Motivated by the observations mentioned, we aim to propose a video summarization method that leverages the saliency information from both static and temporal domains of the video. The static information is extracted from the first frame of the video, while the temporal information is obtained through temporal averaging of the entire video. By combining these two sources of information, the proposed saliency extraction algorithm integrated in the method effectively captures salient information from both static and dynamic cues present in the video. This approach is important because relying solely on a single static image may not always provide complete information, especially when it comes to differentiating between live faces and presentation attacks. Dynamic cues, such as eye blinking, facial expressions, and head movements (e.g., tilting, turning, or nodding), play a crucial role in distinguishing genuine faces from spoofing attempts \cite{wang2021silicone}. Therefore, it is necessary to combine complementary information from the static and temporal domains into a single representative image. This summarization process provides a more compact and informative description of the object under consideration compared to any individual frame from the two sources. By incorporating both static and temporal cues, the proposed method aims to enhance the performance of face presentation attack detection by capturing essential spatio-temporal features that contribute to more accurate and robust results.\\
\indent Our proposed video summarization method has three main advantages: (1) Unlike previous video-based methods \cite{muhammad2022self, muhammad2022adaptive, muhammad2019face} that involve choosing different sampling rates or segments to optimize performance, the proposed method adopts a single-image-based approach for deep learning models. This approach effectively combines spatiotemporal variations present in the videos, providing a more compact and informative representation. (2) Comparing to the scenario of simple frame averaging, the proposed saliency-based fusion is less affected by noise and retains primary structural information, as demonstrated in the fused image in Figure 1. (3) By summarizing the video into a single representative image, the method offers computational efficiency and reduces the memory requirements for deep learning models. This advantage is particularly beneficial in scenarios where accessing and processing every individual frame in a video sequence may be unnecessary or impractical. In summary, the key contributions of the proposed video summarization method can be summarized as follows: 
\begin{enumerate}
 \item This study pioneers the use of saliency to summarize an entire video into a single image specifically for the purpose of face anti-spoofing.
\item  In contrast to optical flow or global motion estimation methods, our approach is computationally faster since it does not involve estimating pixel-level motion vectors between frames. This computational efficiency makes our method more practical and scalable for real-world applications.
\item Through the use of a CNN-RNN network, we evaluate the effectiveness of our approach on five different datasets. The results demonstrate promising generalization ability, and we achieve new state-of-the-art performance in face anti-spoofing tasks.
\end{enumerate}
\begin{figure*}[!t]
\centering
\includegraphics[width=0.85\textwidth]{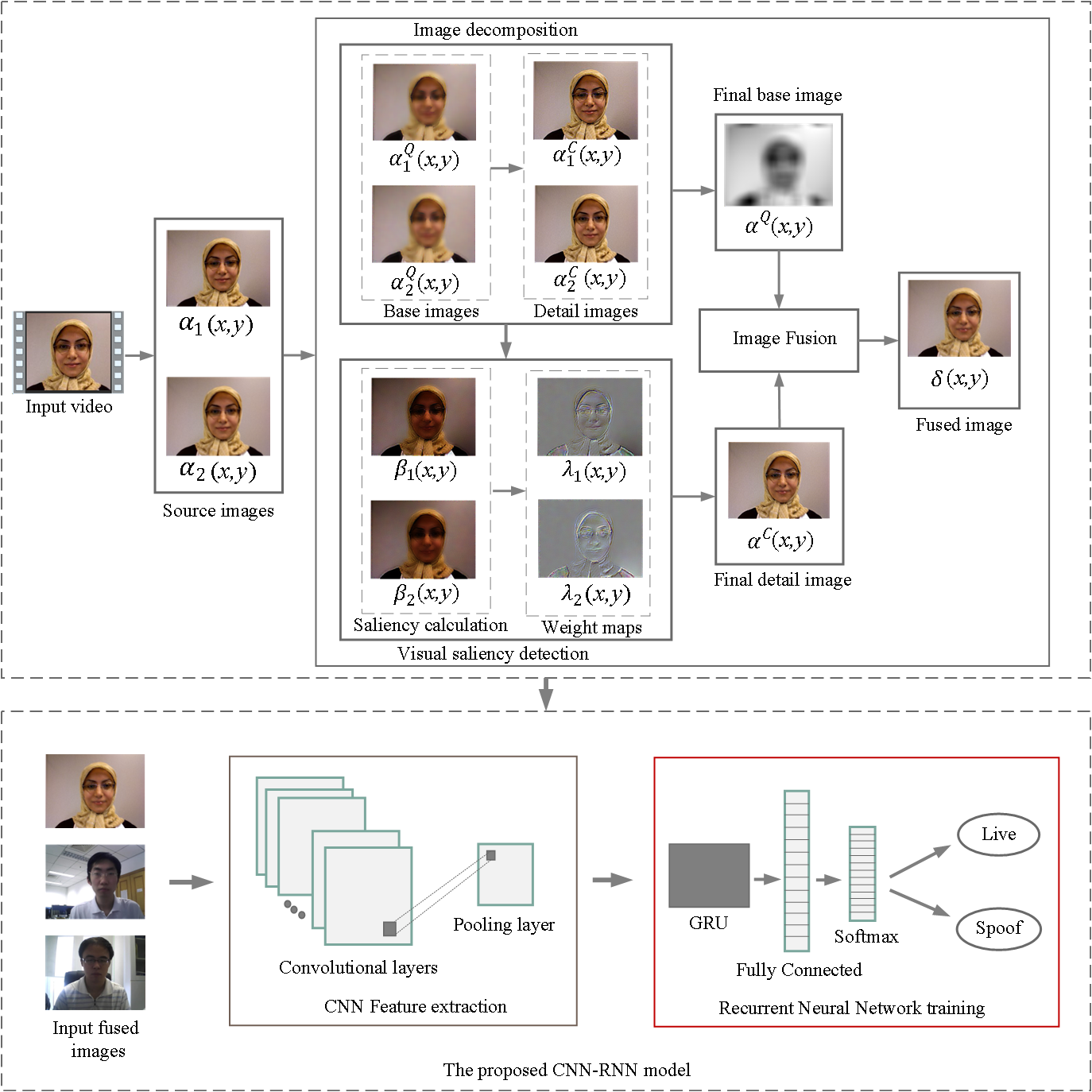}
\vspace{10pt}
\caption{Schematic diagram of the proposed video-based face anti-spoofing countermeasure.}
\end{figure*}
\section{METHODOLOGY}
The proposed approach consists of two main components: (1) a three-stage video summarization model, and (2) a joint CNN-RNN network for face presentation attack detection (PAD). The video summarization model comprises three essential steps: image decomposition, visual saliency detection, and image reconstruction. In the first step, image decomposition is achieved using the Wiener filter \cite{lim1990two} and the Laplacian filter (also known as the Laplacian of Gaussian filter) \cite{burt1987laplacian} to obtain the base and detail images of the source images. This decomposition process enhances the representation of essential information in the frames. Next, visual saliency maps are computed by taking the difference of the Wiener and Laplacian filtering outputs. These saliency maps identify the most salient regions in the images, which are crucial for capturing dynamic cues and temporal variations. In the final stage of the video summarization model, fusion is performed using the final base and detail images. The fusion process combines the complementary information from the static and temporal domains, resulting in a single representative image that effectively summarizes the entire video. Subsequently, the fused images are utilized as inputs for the CNN-RNN network, which aims to detect real and attack images. The CNN-RNN architecture enables the model to effectively capture spatio-temporal features, providing enhanced discrimination between genuine and spoofed faces. Figure 2 visually summarizes our approach, showing the integration of our proposed video summarization method with a CNN-RNN network. Specifically, we employ the DenseNet-201 CNN model \cite{huang2017densely} and a GRU-based RNN model \cite{cho2014learning} to address face presentation attack detection.
\begin{figure}
\centering
\includegraphics[width=0.85\textwidth]{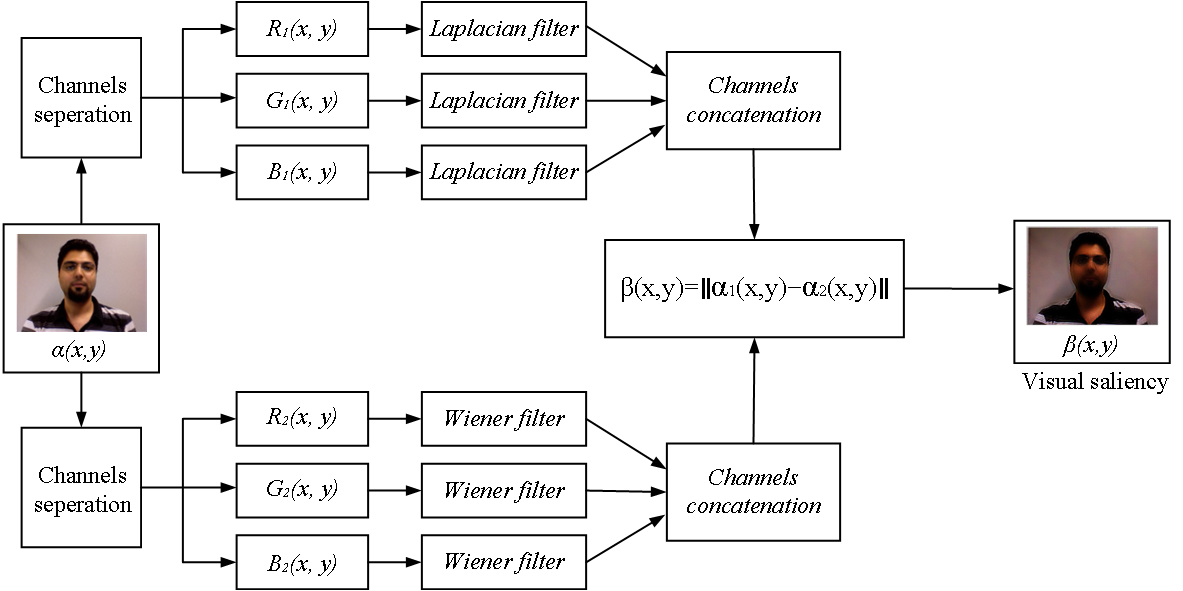}
\vspace{10pt}
\caption{General visualization of the color-based visual saliency estimation.}
\end{figure}
\subsection{Two-scale image decomposition}
In the proposed approach, two co-registered source images, denoted as $\alpha_{1}(x, y)$ and $\alpha_{2}(x, y)$, are used. The first source image is the first frame of the video, while the second source image is obtained through temporal averaging. After obtaining the two source images from the video, an image decomposition process is applied, employing the Laplacian and Wiener filter techniques to transform them into their respective base images, denoted as $\alpha^{Q}_{1} (x, y)$ and $\alpha^{Q}_{2} (x, y)$. In particular, both the Laplacian and the Wiener filter are individually applied to each color channel of the source images. The Laplacian filter is defined as: 
\begin{equation}
\alpha^{Q}_{1} (x, y) = \nabla^2 [G(x, y) \cdot I(x, y)],
\end{equation}
where $\nabla^2$ represents the Laplacian operator. $G(x,y)$ is a 2D Gaussian kernel. $I(x,y)$ is the original color channel. Similarly, the Wiener filter is applied as:
\begin{equation}
\alpha^{Q}_{2} (x, y) = \mathcal{F}^{-1}\left\{H(f_x, f_y) \cdot \mathcal{F}\left[I_n(x, y)\right]\right\},
\end{equation}
where $\mathcal{F}$ represents the Fourier transform. $H(f_x, f_y)$ is the frequency response of the Wiener filter, which is implemented to minimize noise while preserving image features. $I_n(x, y)$ is the noisy color channel (original color channel with added Gaussian noise). By doing so, the detail images are obtained by subtracting the base images from the source images: 
\begin{equation}
\alpha^{C}_{1} (x, y) = \alpha_{1} (x, y) - \alpha^{Q}_{1} (x, y),
\end{equation}
\begin{equation}
\alpha^{C}_{2} (x, y) = \alpha_{2} (x, y) - \alpha^{Q}_{2} (x, y),
\end{equation}
where $\alpha^{C}_{1}$ and $\alpha^{C}_{2}$ are the detail images. The base images contain the essential structural information of the images, representing the low-frequency components, whereas the detail images capture the finer details and high-frequency components of the images. This representative image is subsequently used for further processing, such as saliency detection and fusion. 

\subsection{Visual saliency detection}
Visual saliency detection refers to the process of identifying the most visually salient regions or objects in an image or a scene \cite{bavirisetti2016two}. The goal of visual saliency detection is to simulate and predict the areas that are likely to capture human attention compared to other regions present in the scene \cite{wang2021silicone}. Since temporal averaging is performed, our attention is naturally drawn to elements that exhibit salient features. These features can include eye blinking, hands movement, head rotation, or abrupt changes in texture or orientation. By focusing on salient regions, we introduce a simple saliency map detection algorithm to extract visual saliency for the purpose of fusion. As shown in Figure 3, the Wiener filter \cite{lim1990two} is employed on each source image to reduce noise while preserving important image details. The Wiener filter estimates the original image from the noisy version by considering the noise characteristics and the local image statistics. This helps to enhance the quality of the source images and reduce noise-induced artifacts.\\
\indent Similarly, the Laplacian filter \cite{burt1987laplacian} is applied to each source image to enhance the edges and highlight regions where the intensity changes abruptly. The Laplacian filter computes the second derivative of the image intensity with respect to position, effectively emphasizing regions with significant changes in pixel values. This step is essential for capturing fine details and high-frequency information, which can be important for detecting presentation attacks.\\
\indent The proposed method takes advantage of both the Wiener and Laplacian filtering schemes to estimate the saliency map of each source image. The difference of the filtering outputs is considered because it highlights regions where the pixel values significantly differ between the two images, which are likely to be visually salient. To obtain the magnitude of differences from the difference image, the L2 norm (Euclidean distance), is used. The L2 norm calculates the magnitude of the difference image, representing the overall saliency representation. As visualized in Figure 3, the saliency map is calculated as follows:
\begin{equation}
\beta(x, y) = \lVert \alpha^{Q}_{1} (x, y) - \alpha^{Q}_{2} (x, y) \rVert,
\end{equation}
where $\lVert \cdot \rVert$ represents the $L_{2}$-norm or Euclidean distance. $\alpha^{Q}_{1} (x, y)$ denotes the output of the Laplacian filter and $\alpha^{Q}_{2} (x, y)$ represents the output of the Wiener filter. This saliency map $\beta$ effectively highlights the most visually salient regions in the image, providing important information for further processing steps, such as image fusion.
\subsection{Weight map construction}
The source images such as $\alpha_{1}(x, y)$ (i.e., the still image) and $\alpha_{2} (x, y)$ (i.e, the temporal averaged image) provide complementary information. For instance, the still image provides detailed information about the object but fails to provide the information related to its temporal behavior. While temporal averaging can reduce noise in the images, it may also reduce the visibility of low-contrast details or fine structures in the scene. If the noise levels are low, the temporal averaging process might lead to an unnecessary reduction in the level of detail and texture in the averaged image. Thus, we need to combine visually significant information from both source images to create a composite image that contains the best features or details from each input image. This can be achieved by assigning high weight to pixels with more information and low weight to pixels with insignificant information. Thus, considering the saliency maps, the estimation of weight maps involves normalizing the saliency values to ensure they are in a valid range, typically between 0 and 1. Then, normalization helps prevent biases and ensures that the weights are appropriately scaled for subsequent processing steps. These weight maps are given by:
\begin{equation}
\omega_{1} (x, y)= \frac{\beta_{1} (x, y)}{\beta_{1} (x, y) + \beta_{2} (x, y)},
\end{equation}
\begin{equation}
\omega_{2} (x, y)= \frac{\beta_{2} (x, y)}{\beta_{1} (x, y) + \beta_{2} (x, y)},
\end{equation}
where $\beta_{1}$ and $\beta_{2}$ are computed according to Eq. 5 from $\alpha_{1}$ and $\alpha_{2}$, respectively. This weighted fusion process ensures that important information from both the static and temporal domains is appropriately preserved. 

\subsection{Detail image fusion}
Since the source images are decomposed into base images and detail images, the purpose of detail image fusion is to enhance the level of detail, sharpness, or texture in the resulting image by multiplying the weight maps $\omega_{1}$ and $\omega_{2}$ with the detail images $\alpha^{C}_{1}$ and $\alpha^{C}_{2}$, respectively:
\begin{equation}
\alpha^{C}(x, y)= \omega_{1}(x, y)\alpha^{C}_{1} (x, y) + \omega_{2}(x, y)\alpha^{C}_{2} (x, y),
\end{equation}
where $\alpha^{C}$ is the final detail image. By applying the weight maps to the corresponding detail images, the fusion process effectively combines the complementary information present in both detail images. 

\subsection{Base image fusion}
The fused base image is obtained by using an average fusion rule: 
\begin{equation}
\alpha^{Q}(x, y)= \frac{1}{2}(\alpha^{Q}_{1}(x, y) + \alpha^{Q}_{2}(x, y)),
\end{equation}
where $\alpha^{Q}$ is the final base image. 
\subsection{Image composition}
The final output image is reconstructed by combining the fused base image with the fused detail image. As shown in Figure 2, the detail image typically contains high-frequency information, such as fine textures, edges, or intricate patterns. On the other hand, the fused base image represents the integrated low-frequency information. The reconstruction process is achieved by adding the fused base image $\alpha^{Q}(x, y)$ to the fused detail image $\alpha^{C}(x, y)$. The formula for the reconstruction is as follows:
\begin{equation}
\alpha^{f}(x, y) = \alpha^{Q}(x, y) + \alpha^{C}(x, y),
\end{equation}
where $\alpha^{f}(x, y)$ represents the final output image. By combining the fused base image and detail image, the reconstruction step ensures that both the low-frequency and high-frequency information are appropriately integrated, resulting in a visually coherent and balanced output image. This final output image serves as the single representative image summarizing the entire video and containing the most important information for face presentation attack detection.
\begin{table*}[t]
\centering
\caption{Performance evaluation using MSU-MFSD (M), Idiap (I), CASIA (C) and OULU-NPU (O) databases. Comparison results are obtained directly from the cited papers.} \label{tab:cap2}
\vspace{8pt} 
\resizebox{11cm}{!}{%
\begin{tabular}{l|r|l|r | l| r | l|r| l}
\hline
 & \multicolumn{2}{c|}{O\&C\&I to M} & \multicolumn{2}{c|}{O\&M\&I to C}  & \multicolumn{2}{c|}{O\&C\&M to I} & \multicolumn{2}{c}{I\&C\&M to O} \\ \hline
  Method     & HTER   & AUC   & HTER   & AUC  & HTER    & AUC    & HTER  & AUC   \\ \hline 
MADDG  \cite{shao2019multi} & 17.69    & 88.06   &  24.50 &  84.51 & 22.19 &   84.99 &  27.89 &   80.02 \\
DAFL  \cite{saha2020domain} & 14.58    & 92.58   &  17.41 &  90.12 & 15.13 &  95.76 &  14.72 &  93.08 \\ 
SSDG-R  \cite{jia2020single} & 7.38    & 97.17   &  10.44 &  95.94 & \underline{11.71} &   \textbf{96.59} &  15.61 &   91.54 \\ 
DR-MD  \cite{wang2020cross} & 17.02   & 90.10 &  19.68 &  87.43 & 20.87 & 86.72 &  25.02 &  81.47 \\
MA-Net  \cite{liu2021face} & 20.80    &  ---  &  25.60 & ---  & 24.70 & ---  &  26.30 &  ---  \\
RFMetaFAS  \cite{shao2020regularized} & 13.89    &  93.98  &  20.27 &  88.16 & 17.30 & 90.48 &  16.45 &  91.16 \\
FAS-DR-BC(MT)  \cite{qin2021meta} & 11.67    & 93.09   &  18.44 &  89.67 & 11.93 & \underline{94.95} &  16.23 &  91.18 \\
ASGS \cite{muhammad2022adaptive}  & 5.91   & \textbf{99.88} &   10.21  & \underline{99.86}  &  45.84 &  76.09  &  13.54  &   \textbf{99.73}  \\
HFN + MP  \cite{cai2022learning} & 5.24 & 97.28 &  \underline{9.11}  &   96.09 &   15.35 &   90.67  & \underline{12.40} &  94.26 \\ 
Cross-ADD \cite{huang2022generalized}  & 11.64   & 95.27 &   17.51  & 89.98  &  15.08 &  91.92  &  14.27  &    93.04  \\ 
FG +HV  \cite{liu2022feature} & 9.17   & 96.92  &  12.47 &  93.47 & 16.29  &  90.11 &  13.58 &  93.55 \\
ADL  \cite{liu2022adversarial} &   5.00   &   97.58    & 10.00 &  96.85 & 12.07 &   94.68  & 13.45  &  94.43 \\ 
           \hline 
Ours &  \textbf {4.51} & 99.17  & \textbf {4.16} & \textbf {99.96} &    \textbf {9.75} & 94.65  &  \textbf{8.47} & 96.80 \\ 
\hline\end{tabular}}
\end{table*}

\begin{table}[h]
\centering
\caption{Combined cross-database evaluation using MSU-MFSD (M), Idiap Replay-Attack (I), CASIA-FASD (C), and OULU-NPU (O) databases. Comparison results are obtained directly from the cited papers.}
\label{tab:combined}
\begin{tabular}{lc|c|c|c}
    \hline
    & \multicolumn{2}{c|}{M\&I to C} & \multicolumn{2}{c}{M\&I to O} \\
    \hline
    Method & HTER(\%) & AUC(\%) & HTER(\%) & AUC(\%) \\
    \hline
    MS-LBP \cite{maatta2011face} & 51.16 & 52.09 & 43.63 & 58.07 \\
    LBP-TOP \cite{de2014face} & 45.27 & 54.88 & 47.26 & 50.21 \\
    Color-LBP \cite{boulkenafet2016face} & 55.17 & 46.89 & 53.31 & 45.16 \\
    IDA \cite{wen2015face} & 45.16 & 58.80 & 54.52 & 42.17 \\
    MADDG \cite{shao2019multi} & 41.02 & 64.33 & 39.35 & 65.10 \\
    SSDG-M \cite{jia2020single} & 31.89 & 71.29 & 36.01 & 66.88 \\
    \hline
    Ours & \textbf{23.88} & \textbf{99.78} & \textbf{16.87} & \textbf{94.05} \\
    \hline
\end{tabular}
\end{table}

\subsection{CNN-RNN network}
Convolutional Neural Networks (CNNs) are well-suited for processing the spatial information. To encode temporal information, we combine the strength of CNN in capturing spatial features and the temporal modeling capabilities of the recurrent neural network (RNN), in partiuclar a Gated Recurrent Unit (GRU) \cite{cho2014learning}. We specifically utilize a 2D (CNN) \cite{huang2017densely} to extract off-the-shelf CNN features from the fused images. Subsequently, the $1920$-dimensional features obtained from the pooling layer of a CNN are used as input for training the GRU model. The GRU layers learn temporal patterns and dependencies in the sequential data, integrating the spatial information by updating its hidden state based on the current input and its previous hidden state. By combining CNN for spatial feature extraction and GRU layers for capturing temporal dependencies, the network can effectively process and analyze sequential or temporal data with spatial structure, leading to improved performance in tasks involving both spatial and temporal information.

\section{Experiments}
To evaluate the performance, OULU-NPU \cite{boulkenafet2017oulu} (denoted as O), Idiap Replay-Attack \cite{chingovska2012effectiveness} (denoted as I), MSU Mobile Face Spoofing \cite{wen2015face} (denoted as M), and CASIA Face Anti-Spoofing (denoted as C) databases are utilized in our work. The Half Total Error Rate (HTER) is reported that provides an overall measure of the error rate in biometric authentication systems, balancing both false acceptances and false rejections. Moreover, a Silicone Mask Face Motion Video Dataset (SMFMVD) \cite{wang2021silicone} is also evaluated to detect the silicone mask faces against counterparts.

\subsection{Implementation details}
All the images are resized to  $224 \times 224$ to extract deep features from the the Densely Connected Convolutional Networks (DenseNet-201) \cite{huang2017densely} model. Data augmentation transformations such as rotation, x-translation, and y-translation are applied to improve the robustness and generalization of the model. For training GRU, the parameters such as the Adam optimizer with the hidden layer dimension of $500$, validation frequency $30$, and learning rate $0.0001$ was used to develop the model. We do not set the fixed epochs because an early stopping function \cite{prechelt2002early} was utilized. The He initializer \cite{he2015delving} is utilized as weight initialization technique that helps to ensure more stable and effective learning. We keep the same parameters for training all the datasets for the repeatability of the results. 
\begin{table}[t]
  \centering
  \caption{The execution time for estimating TSS method, optical flow, and ours.}
  \label{tab:cap}
  \begin{minipage}{0.9\textwidth}
    \begin{tabular}{cccc}
      \hline
      Dataset & Optical flow \cite{horn1981determining} & TSS method \cite{muhammad2022self} & Ours \\
      \hline
      CASIA-FASD & 36.1 & 12.4 & 2.1 \\
      REPLAY-ATTACK & 31.8 & 7.2 & 1.6 \\
      \hline
    \end{tabular}
  \end{minipage}
\end{table}

\begin{table}[h]
\centering
\caption{The performance evaluation in terms of intra-dataset. Comparison results are obtained from the original paper \cite{wang2021silicone}.}\label{tbl6}
\begin{tabular}{llllll}
\toprule
 & \multicolumn{4}{c}{SMFMVD Dataset} \\
\cmidrule{2-5}
& APCER(\%) & BPCER(\%) & ACER(\%) & EER(\%)  \\
\midrule
CDCN++ \cite{yu2020searching}  & 10.0   & 36.0 & 23.0   & 21.5 \\
IDA \cite{wen2015face}  & 35.0 & 4.00 & 19.5  & 14.1  \\
Videolet  \cite{wang2021silicone}  & 4.0    & 73.0  & 38.5   & 22.1  \\
CT  \cite{boulkenafet2016face}  & 9.0    & 15.0 & 12.0   & 12.3 \\
Ref. \cite{wang2020deep}  & 8.0   & 51.0 & 29.5 & 34.1 \\
VSFM \cite{wang2021silicone}  & 4.0   & 2.0 & 3.0   & 1.1  \\
ASGS \cite{muhammad2022adaptive}  & 3.7   & 0.1  & 2.8   & 1.6  \\
\midrule
Ours  &  \textbf{3.5}  & \textbf{0.1}  & \textbf{2.3}   & \textbf{0.7}  \\
\bottomrule
\end{tabular}
\end{table}
\begin{figure}
\centering
\includegraphics[width=0.5\textwidth]{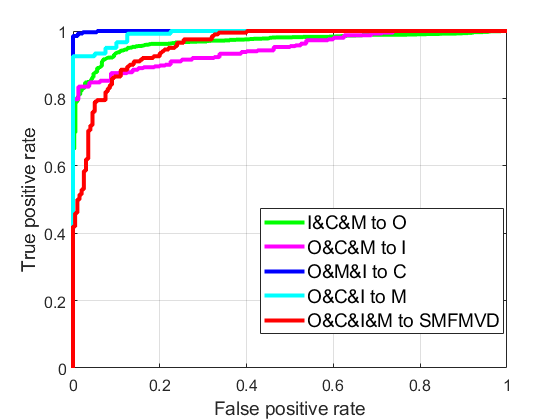}
\vspace{10pt}
\caption{The Receiver Operating Characteristics (ROC) curves of cross-dataset testing.}
\end{figure}

\subsection{Comparison against the state-of-the-art methods}
To evaluate the effectiveness of the proposed approach, we compare the performance with several state-of-the-art deep learning models. The leave-one-out (LOO) strategy is adopted where three datasets are randomly selected for training and one dataset is left out for testing. In contrast to previous approaches \cite{cai2022learning, yu2020fas, saha2020domain, shao2019multi, muhammad2022adaptive, cai2022learning, liu2022adversarial}, our proposed method achieve the best HTER on all the four domain generalization test sets. By emphasizing salient information, we hypothesize that the summarization process can provide a more focused and informative representation that captures the essence of the video. Therefore, this can enhance the model's ability to generalize from the summarized representation to unseen videos. Considering a limited source domain (i.e.,  two source domains training) as shown in Table 2, one can see that our proposed method continues to compete the previous approaches by a fair margin. We also use another evaluation metric such as area under the curve (AUC) that provides a comprehensive evaluation of the model's performance by capturing the trade-off between sensitivity (true positive rate) and specificity (1 - false positive rate). It can be observed from Table 1 and 2 that the proposed method attains more than $90\%$ AUC on all the datasets.\\
\indent Moreover, we compare the computational time of our video summarization method with other motion estimation techniques such as optical flow \cite{horn1981determining} and TSS learning \cite{muhammad2022self}. The comparison is performed by taking a random video of CASIA and Replay-Attack datasets. Table 3 shows the time (in terms of seconds) required for performing the estimation by considering all frames of video.  All the experiments are conducted in a MATLAB environment by using a workstation with 3.5 GHz Intel Core i7-5930k and 64 GB RAM. The results demonstrate that the computational time is far superior than the methods based on global motion (TSS) and the optical flow algorithm. Since we evaluated our proposed method on those datasets that do not focus on silicone mask attack, it would remain unclear how the video summrization method can contribute to silicone mask attacks. Therefore, we conduct experiments on a silicone mask face motion video dataset \cite{wang2021silicone} and report the performance in Table 4. The experimental results show that the performance of the proposed method is superior over the state-of-the-art methods under intra-dataset test environment.\\
\indent For further analysis, the ROC curves are created in Fig. 4  by plotting the true positive rate (TPR) on the y-axis and the false positive rate (FPR) on the x-axis. The ROC curves can be seen on the top-left corner, indicating higher true positive rates and lower false positive rates to show better discrimination ability. We also show the ROC curve when the model trained on four datasets (i.e. O\&M\&I\&C) and evaluated on SMFMVD dataset. The model's performance on the SMFMVD dataset indicates its generalization capability and competitive performance even on unseen silicone mask images.

\section{Conclusions}
The paper presents a novel video summarization method tailored for the face anti-spoofing task, offering a concise and informative representation of the video content. By prioritizing salient information, the proposed summarization approach enables the model to focus on the most relevant regions efficiently. The generation of a single image from each video reduces computational load and enhances the model's generalization ability. Furthermore, the study demonstrates that the proposed video summarization method complements a simple deep learning architecture (CNN-RNN) and achieves state-of-the-art performance on five face anti-spoofing datasets. This indicates the effectiveness of the approach in capturing essential spatio-temporal features and improving the accuracy of the face anti-spoofing system. However, the paper acknowledges a potential limitation of the approach: it might not capture all fine details, especially in longer videos. Future research directions could focus on exploring new approaches and improvements to overcome this limitation and develop more effective video summarization methods for longer videos. 


\section*{Acknowledgment}
This work is financially supported by ‘Understanding speech and scene with ears and eyes (USSEE)’’ (project number 345791). The first author also acknowledges the support of the Ella and Georg Ehrnrooth foundation.

\begin{appendices}




\end{appendices}


\bibliography{sn-bibliography}

 %
 
\end{document}